# Braille to Text Translation for Bengali Language: A Geometric Approach

Minhas Kamal
Institute of Information Technology
University of Dhaka
Dhaka, Bangladesh
Email: bsse0509@iit.du.ac.bd

Dr. Amin Ahsan Ali and Dr. Muhammad Asif Hossain Khan
Department of Computer Science & Engineering
University of Dhaka
Dhaka, Bangladesh
Email: {aminali, asif}@du.ac.bd

Dr. Mohammad Shoyaib
Institute of Information Technology
University of Dhaka
Dhaka, Bangladesh
Email: shoyaib@du.ac.bd

***Abstract*— Braille is the only system to visually impaired people for reading and writing. However, general people cannot read Braille. So, teachers and relatives find it hard to assist them with learning. Almost every major language has software solutions for this translation purpose. However, in Bengali there is an absence of this useful tool. Here, we propose Braille to Text Translator, which takes image of these tactile alphabets, and translates them to plain text. Image deterioration, scan-time page rotation, and braille dot deformation are the principal issues in this scheme. All of these challenges are directly checked using special image processing and geometric structure analysis. The technique yields 97.25% accuracy in recognizing Braille characters.**

## I. INTRODUCTION

In Braille writing system, dots are embossed on paper. A group of dots represent certain character of any specific language, which is perceivable by the finger tip. Optical Braille Recognition process takes image of these tactile alphabets, and translates them to plain text.

Visually impaired people have all the abilities that any normal person need, to contribute to the society. The only problem they face is not being able to visually perceive the world. As most of our learning procedures are largely dependent on the eye, a big portion of this human potential is not getting the room for manifestation. So rather becoming an asset, they turn into a burden to the community [1].

This problem was intensely severe till the early 1800's, when Braille Writing system came into existence. The base process was developed by the French Army for passing message in the dark. The system was then matured by a young boy named Braille, who accidently lost his vision at an early age [2]. He simplified the writing and reading process, and as a result the technique became very popular among visually impaired people [3]. Further refinement to the procedure shaped it for mass use. Now, almost every major language in this world has its own Braille format [4].

Braille is a specialized writing system for visually impaired people. Here raised dots on embossed paper are used as tactile alphabet [5] [6]. For example: Figure 1 represents Bengali alphabet- 'আ'.

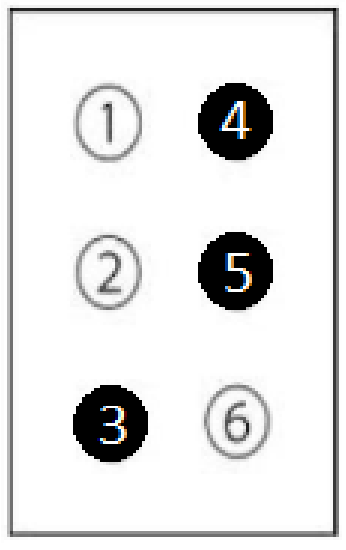

Figure 1: A Braille Cell with 6 dots Representing- 'আ'

In exam, students with visual impairment write using braille. However, it is quite tough to find a competent examiner who also reads braille. So, most of the time a human translator is used during the evaluation process which is both tiring, slow, as well as prone to human error. In this situation, a Braille translator would give a huge edge to the examination system. The translator will also help parents while teaching their children.

Standing in this era of modern science we wish to get every bit of information in digital form. Therefore, there are some very well-known commercially available text-to-braille translators such as- Win Braille [7], Cipher Braille Translator [7] and Braille Master. These tools demonstrate high performance both in computational time and accuracy. However, braille-to-text translators face much more complications than theirs counterpart. And so, both effectiveness and precision suffer during Braille Character Recognition. Moreover, very few researches have been performed over Bengali Braille to text translation. And, to the best of our knowledge, there is also no single open source tool available for this translation purpose.

The goal of this paper is to present a brief description of the current state of art Braille processing techniques, and then propose a convenient system for translating Bengali Braille writing to text. Braille written embossed paper is scanned by a typical flatbed scanner. The scanned image is generally tilted and contains lots of noise (Figure 4b). In order to address these issues, different image processing procedures were followed, like- image enhancement, noise reduction, and rotation invariant geometric structure recognition. Braille dot reformation has been dealt using geometric pattern recognition techniques. And then translate it to text.

## II. LITERATURE REVIEW

Optical Braille Recognition (OBR) System consists of three major modules- Image Acquisition and Preprocessing, Dot Localization and Extraction, and ultimately Braille Character Detection and Classification. Several attempts had been made to translate scanned image of Embossed Braille Writing to text.

In 1988 an algorithm called Lectobraille [8] was devised by Dubus et al. The algorithm interpreted Braille Writing into print by utilizing filtering and morphological operation followed by an adaptive recognition method.

Mennens with his team formulated an optical recognition system in 1994 for reading both single and double sided Braille [9]. The system could recognize image of Braille writing scanned by any commercial scanner. Their technique does provide great accuracy, but with an assumption of perfect Braille Dot alignment which hardly occurs in factual world. The next year, Blenkhorn et al. proposed a table driven method for supporting braille translation process of a wide varieties of languages [10].

In 1999, Ng and his team used image enhancement, segmentation, and boundary detection to approach the problem [11]. They were able to handle page rotation, but no mention was made to cope with braille character deformation.

In the year of 2001, Murray and Dais designed a handheld device that could optically read and translate Braille Writing [12][13]. The method of capturing image with a portable device provided a much better imaging than that of a flatbed scanner. However, it is difficult to keep the device vertical with respect to the Braille line.

In 2004, Wong et al. used a simple image processing algorithm to detect half character, and probabilistic neural network to recognize half character [14]. The algorithm processes the image one row at a time, and also preserves the layout of the original document.

Mousa with his team suggested a system in 2013 for recognizing image of braille writing scanned by a flatbed scanner [15]. They have used Average Filtering and Contrast Enhancement in the Image Enhancement step. They perform dilation after inversing the image, and then binarize the image for dot segmentation. Lastly, Braille Dot alignment is detected, and characters are recognized. The method deals with page rotation too.

## III. PROPOSED METHODOLOGY

Paper sheets of Bengali Braille writing are scanned by a regular scanner, and stored as image. These images are individually processed to finally generate text. The whole process can be segmented into three steps-

1. Pre-Processing
2. Geometric Pattern Recognition
3. Translation

In the scanner educed raw image, Braille dots usually have very small gradient difference from the background. So they are almost indistinguishable. The Image Pre-Processing step improves contrast, as well as reduces noise from the image. Therefore, the dots become much more differentiable and clear. In Geometric Pattern Recognition period, braille dots are identified, plus their orientation is detected. Lastly, Translation stage traces each individual braille character, and then translates them to a specific language, here Bengali.

**Algorithm- Braille to Text Translation**

**Input**: Scanned Image

**Output**: Text

```
1:  image := bi-histogram-equalization(image)
2:  image := median-filter(image)
3:  image := quantization(image, otsu-threshold)
4:  image := morphological-operation(image)
5:  connected-components[] = segmentation(image)
6:  braille-dots[] = filtration(connected-components)
7:  braille-structure = pattern-recognition(braille-dots)
8:  braille-code = dot-detection(image, braille- structure)
9:  text = translation(braille-code)
```

The subsequent paragraphs will walk through each of the steps in details.

*1. Pre-Processing*

As discussed earlier, Braille dots cannot be dissociated easily from a raw scanned image. So, image is first fed straight to Bi-Histogram Equalization [16]. The method preserves image brightness but enhances contrast. Consequently, braille dots get a higher gradient. However, the method also introduces a lot of noise to the image, mostly salt-and-pepper noise. Hence, the image is passed through a Median Filter [17], a spatial nonlinear filtering technique that is most suited for this type of job. The procedure removes maximum noise, causing minimum effect over the Braille dots.

Then, the image is quantized to binary utilizing Otsu Thresholding [18] [19]. This technique is a statistical process that automatically calculates the optimum threshold to separate a bi-modal probability distribution. From heuristics, most of the pixel's grey-level fall into the upper range, which affects the thresholding procedure severely. As a result, the image histogram is calculated for the lower grey-scale. The histogram is passed to Otsu Thresholding, and depending on threshold the image is quantized to binary. Now, every individual dot should be a single connected component. Yet, many dots are found divided into two or more parts. For joining these dot-parts together Morphological Closing [20] [22] function is performed. The function includes a Dilation Operation followed by an Erosion Operation.

At the end of Pre-Processing, Braille dots appear clean and sharp, ready for Pattern Recognition.

*2. Geometric Pattern Recognition*

For recognizing Braille characters each of their positions must be detected individually. And so the Geometric Braille Writing Structure needs to be analyzed. The structure comprises of following properties-

i. Start Point of Writing- $P_0$ $(x_0, y_0)$
ii. Writing Rotation- $\theta_B$
iii. Braille Character Size- $S_c$ $(w_c, h_c)$
*iv.* Distance between Characters- $d_c$
v. Distance between Lines- $d_L$
vi. Number of Characters in Each Line- $n_L$
vii. Number of Lines- $L_T$

Figure 2 provides a visual understanding for the properties of Geometric Braille Writing Structure.

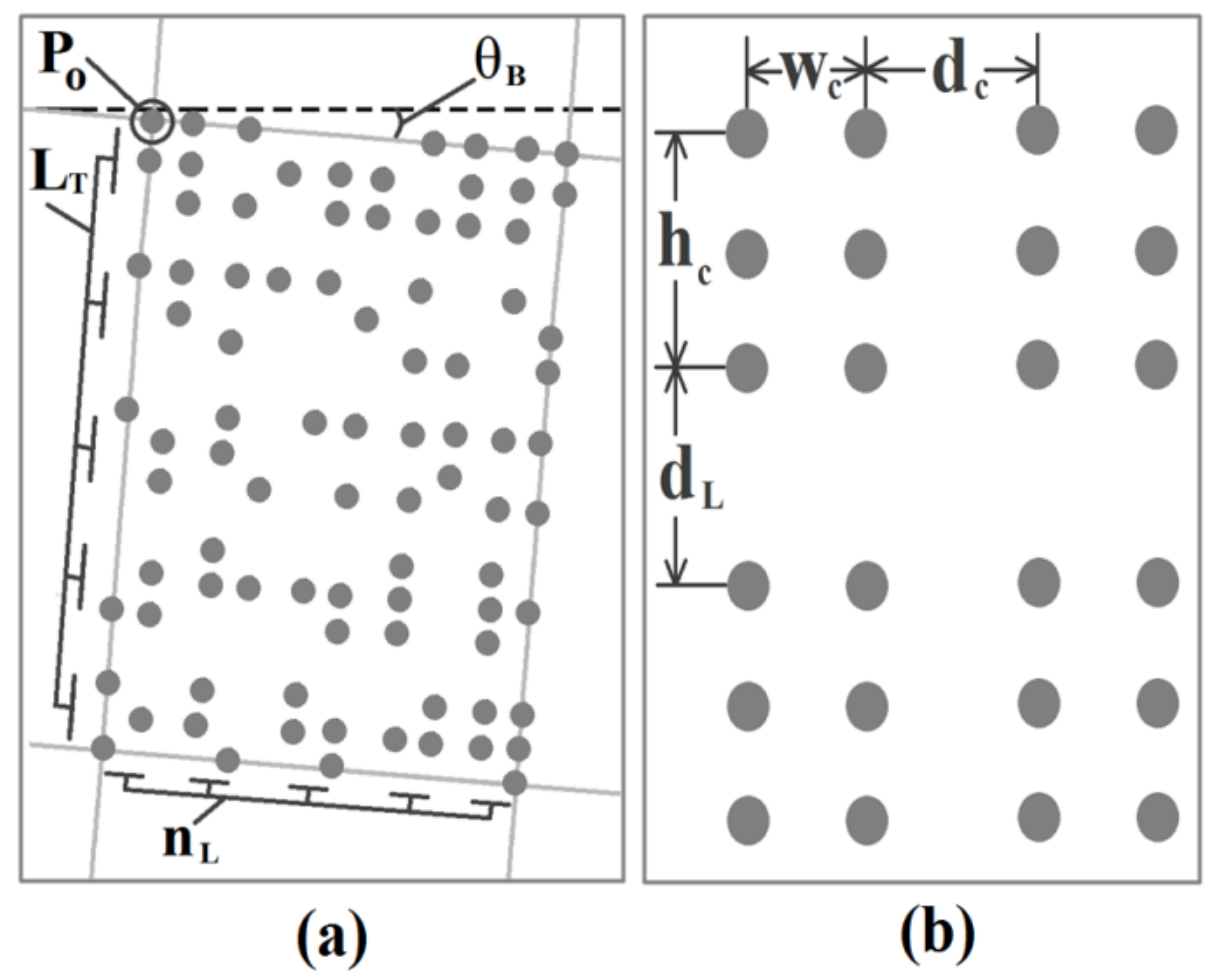

Figure 2: Properties of Geometric Braille Writing Structure, (a) Shows Start Point of Writing- $P_0$, Writing Rotation- $\theta_B$, Number of Characters in Each Line- $n_L$, and Number of Lines- $L_T$, (b) Shows Braille Character Width- $w_c$ and Height- $h_c$, Distance between Characters- $d_c$, and Distance between Lines- $d_L$

To calculate these properties three consequent processes are performed which are discussed in detail below-

*a. Braille Point Detection*

Connected components [22] are extracted of the processed binary image from Pre-Processing stage. Components having a similar width and height are considered as circle shapes. So, to qualify as a circle following condition (1) must satisfy-

$$|\,\omega - \eta| \leq \min(\omega, \eta) \tag{1}$$

Here,

$\omega$ = connected component width

$\eta$ = connected component height

Diameter of each circle shaped connected component is approximated by the mean value of the width and height as shown in equation (2)-

$$\delta = \frac{\omega + \eta}{2} \tag{2}$$

Here,

$\delta$ = diameter of circle shaped connected component

The median value of the diameters is chosen as Standard Braille Dot Diameter ($\delta_S$). All the braille dot diameters are then compared with $\delta_S$, and only the matched dots are used for farther pattern recognition steps. A dot is considered a match, if it falls in the heuristically supported range (3)-

$$\left[\frac{2}{3}\,\delta_s,\ \frac{4}{3}\,\delta_s\right] \quad (3)$$

These matched dots are represented as Braille Points, shown in Figure 2a. Each Braille Point (P) is spread over a 2D matrix having a horizontal co-ordinate (x) and a vertical co-ordinate (y). These co-ordinates are derived during the connected component extraction phase.

*b. Margin Line Determination*

First, Braille Points at the upper most layer are identified. The process starts by sectioning the 2D matrix vertically into stripes, where breadth of each stripe is equal to the $\delta_s$. In each stripe, point having the upper most position is picked. These points are called the Upper Margin Points (Figure 2b).

Now, a thick line of the width of $\delta_s$ is drawn that passes through maximum number of Upper Margin Points. This line is called the Upper Margin Line (Figure 2c).

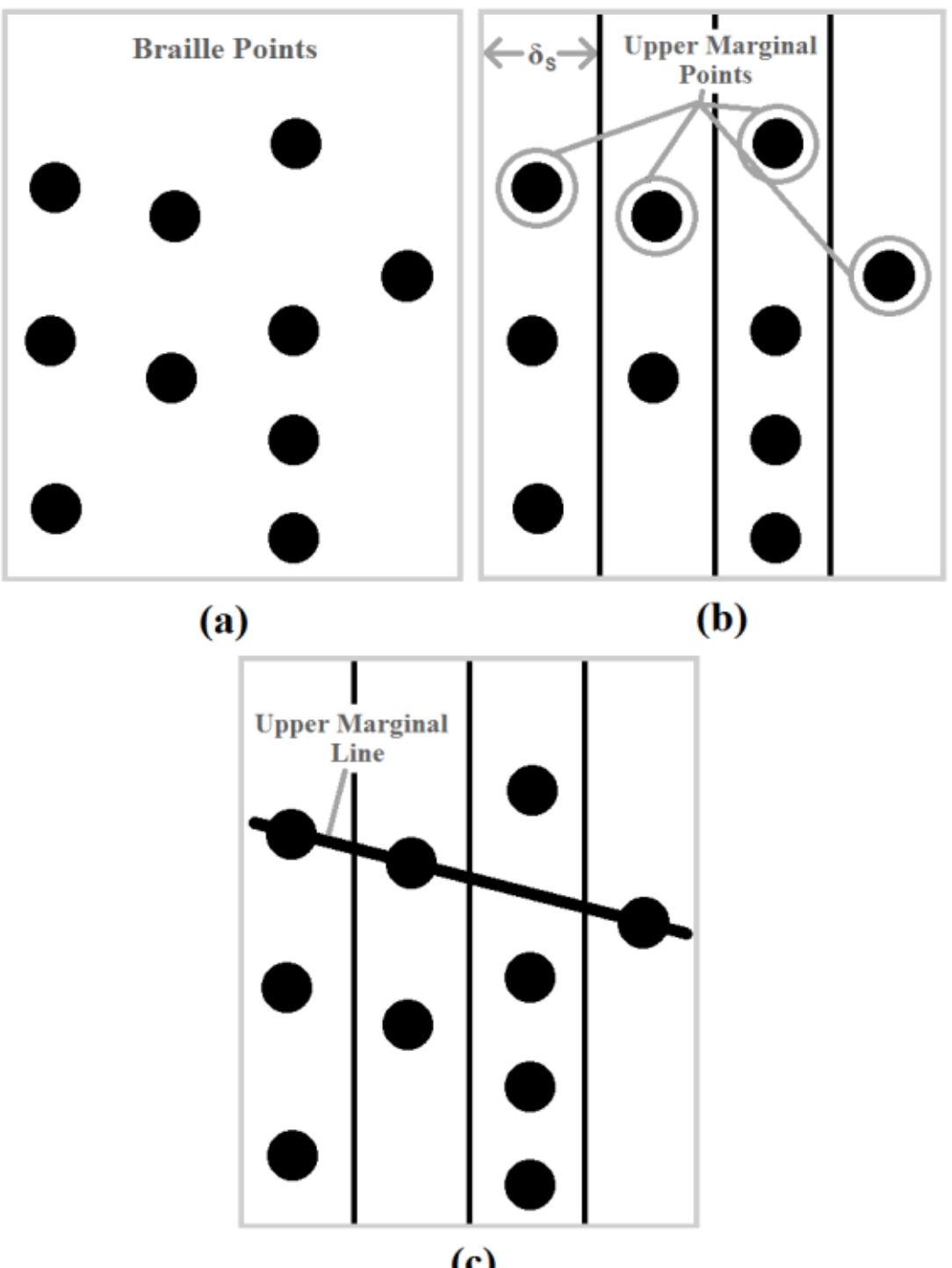

Figure 3: Determining Upper Margin Line, (a) Braille Points Spread over A 2D Matrix, (b) Detecting Upper Margin Points, (c) Drawing Upper Margin Line

The same algorithm is applied for the remaining sides of the matrix; and Lower, Left, and Right Margin Lines are acquired. The intersection point of Upper and Left Margin Line is the Start Point of Writing- ($P_0$). Mean of the slopes of the Margin Lines is Writing Rotation ($\theta_B$).

*c. Peer Distance Computation*

For each adjacent Upper Margin Point pair, horizontal distances are computed and their frequencies are arranged in a Frequency Distribution Vector. Then the vector is smoothed by sliding an N-Point Running Mean [21]. Second derivative of the Smooth Frequency Distribution yields two values- the smaller value corresponds to Braille Character width, the larger value corresponds to distance between Characters.

Same procedure is followed for Left Margin Point pairs, and Braille character height and distance between lines are accomplished.

Text Width can be determined by computing the distance between Left Margin Line and Right Margin Line. Now, number of characters in each line can be estimated from following equation (4)-

$$W_T = n_L * w_c + (n_L - 1) * d_C$$

$$\Rightarrow n_L = \frac{W_T + d_C}{w_c + d_C} \quad (4)$$

Here,

$W_T$ = text width

$w_c$ = Braille character width

$n_L$ = number of characters in each line

$d_C$ = distance between characters

The same way as above text height can be calculated and used in equation (5) for estimating number of lines-

$$L_T = \frac{H_T + d_L}{h_c + d_L} \quad (5)$$

Here,

$L_T$ = number of lines

$H_T$ = text height

$d_L$ = distance between lines

$h_C$ = Braille character height

*3. Translation*

A grid structure is created to en-box each Braille Character depending on the Geometric Braille Writing Structure achieved from previous Geometric Pattern Recognition phase. This grid structure is projected on the binary image attained from Pre-Processing phase. For each grid-cell, a six bit code called Binary Braille Code is generated depending on the presence of black pixels. Like, Figure 1 would present this Binary Braille Code- 001110.

Each of these Binary Braille Codes has a predefined one-to-one correspondence with Bengali Characters (see Table I). For example, the code 001110 denotes the Bengali Character- 'আ'. Thus, all the binary codes are directly mapped and converted to text. By altering this table the technique can work nearly for any language.

TABLE I

ONE-TO-ONE MAPPING OF BINARY BRAILLE CODES AND CORRESPONDING BENGALI CHARACTERS

| **Binary Braille Codes** | **Bengali Charact-ers** | **Binary Braille Codes** | **Bengali Charact-ers** |
|---|---|---|---|
| **Bengali Vowels** | | | |
| 100 000 | অ | 001 110 | আ |
| 010 100 | ই | 001 010 | ঈ |
| 101 001 | উ | 110 011 | ঊ |
| 000 010 | ঋ | 100 010 | এ |
| 001 100 | ঐ | 101 010 | ও |
| 010 101 | ঔ | | |
| **Bengali Consonants** | | | |
| 101 000 | ক | 101 101 | খ |
| 110 110 | গ | 110 001 | ঘ |
| 001 101 | ঙ | 100 100 | চ |
| 100 001 | ছ | 010 110 | জ |
| 101 011 | ঝ | 010 010 | ঞ |
| 011 111 | ট | 010 111 | ঠ |
| 110 101 | ড | 111 111 | ঢ |
| 001 111 | ণ | 011 110 | ত |
| 100 111 | থ | 100 110 | দ |
| 011 101 | ধ | 101 110 | ন |
| 111 100 | প | 011 010 | ফ |
| 110 000 | ব | 111 001 | ভ |
| 101 100 | ম | 101 111 | য |
| 111 010 | র | 111 000 | ল |
| 100 101 | শ | 111 101 | ষ |
| 011 100 | স | 110 010 | হ |
| 111 110 | ক্ষ | 100 011 | জ্ঞ |
| 110 111 | ড় | 111 011 | ঢ় |
| 010 001 | য় | 000 010 | □ |
| 000 100 | ্ | 000 011 | ং |
| 000 001 | ঃ | 001 000 | ঁ |
| **Punctuation Marks** | | | |
| 010 011 | । | 010 000 | , |
| 011 000 | ; | 010 010 | : |
| 011 001 | ? | 011 010 | ! |
| 001 001 | - | | |

## IV. EXPERIMENTAL ANALYSIS

### A. Dataset Description & Implementation Details

The dataset was prepared by scanning embossed paper of braille writing with a commercially available general flatbed scanner. The braille papers were written by visually impaired volunteers. Each page has the capacity to contain 27 lines and each line 28 characters. In total there are 728 characters. The reader should note that the volunteers were given the text to be written, but no other instructions were provided regarding the number of lines to be written in each page, or the line spacing etc. Their writing also contained misspelled words, which were embraced in the result too.

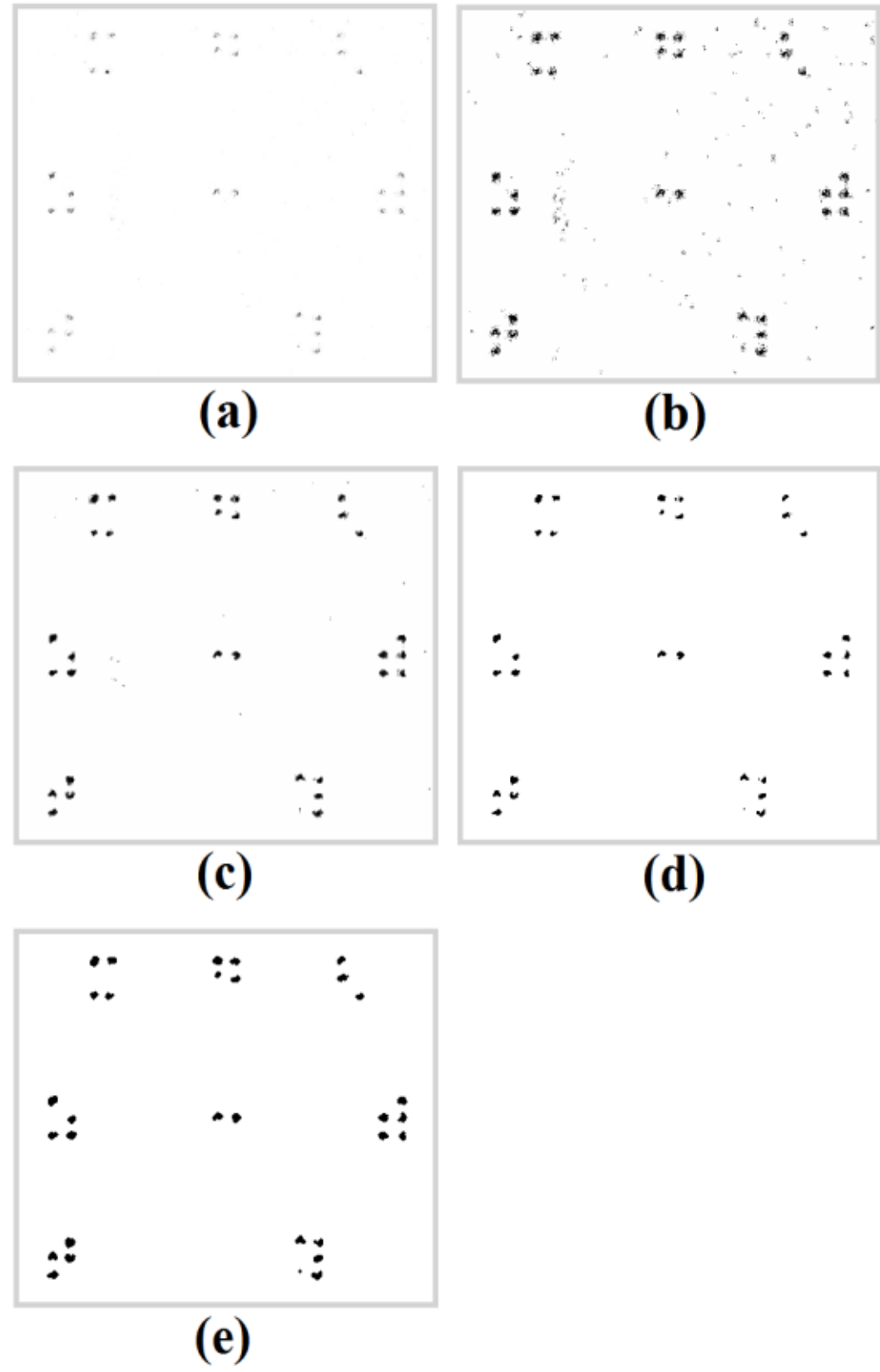

Figure 4: Improving Braille Dots through Image Processing in Pre-Processing Phase, (a) Raw Scanned Image of Braille Writing, (b) Result after Applying Image Enhancement Algorithm- Bi-Histogram Equalization, (c) After Applying Noise Reduction Technique- Mean Filter, (d) Binarizing Image Using- Otsu Thresholding, (e) Running Morphological Operation for Braille Dots' Shape Development

The scanned images contain very faint existence of braille dots over the white paper. However introduces lots of noise; the image is also rotated. So, the Pre-Processing phase acts on these faint dots and improve their appearance. Figure 3 demonstrates step-by-step improvement of braille dots throughout the whole process.

In the first step of Geometric Pattern Recognition phase, dots having higher probability of being actual Braille Dots are chosen as Braille Points and these points are used for subsequent pattern recognition steps (Figure 4a). Authentic Braille Dots are very important as presence of too much noise completely ruins the whole process of automated geometric configuration detection and produces garbage result.

Depending on Geometric Braille Writing Structure a grid construction is formed (Figure 4c), which is then projected over the Pre-Processed Image (Figure 3e) as shown in Figure 4d. Each cell represents one Braille Character.

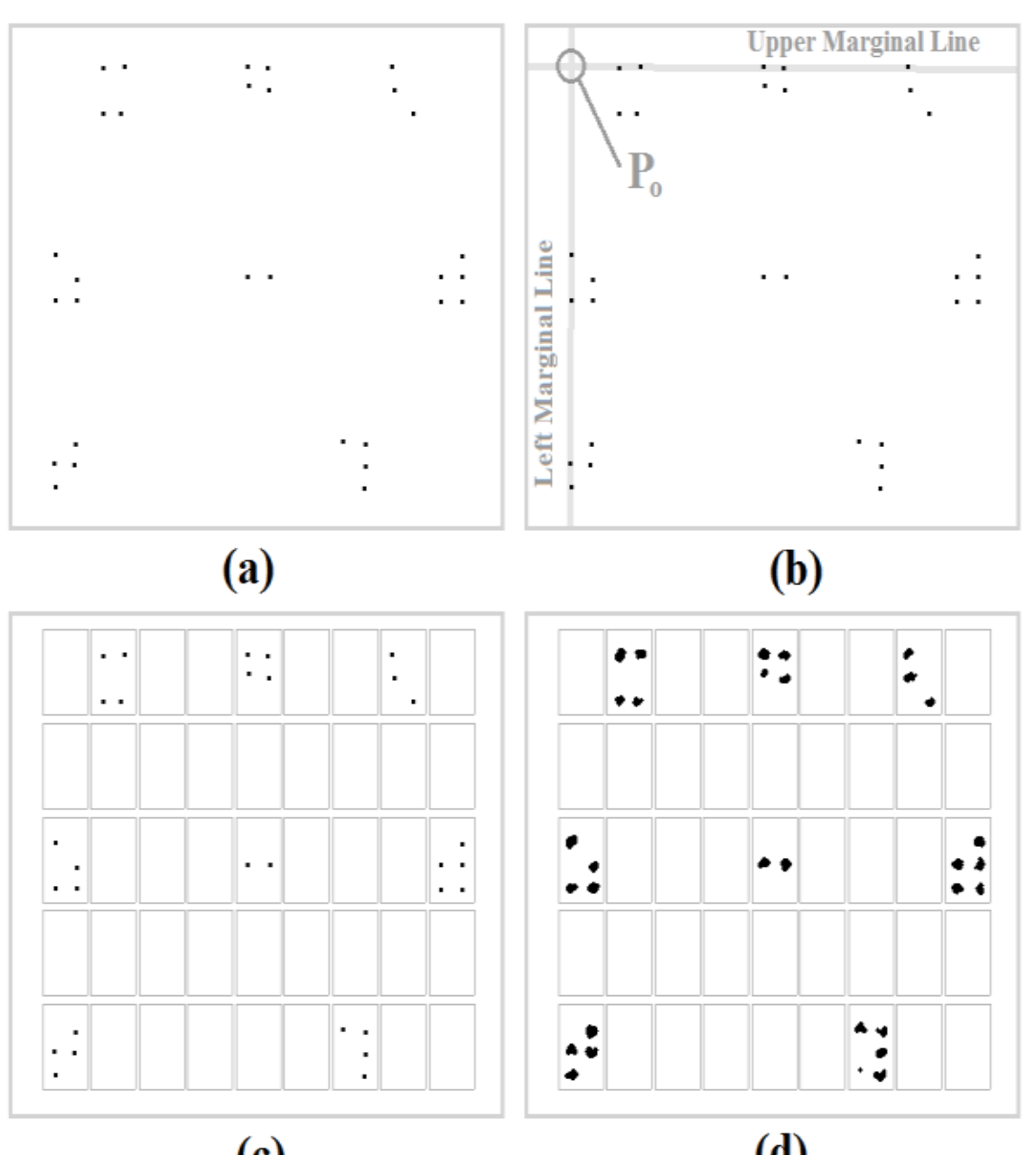

Figure 5: Detecting Braille Dot Alignment in Geometric Pattern Recognition Phase, (a) Identifying Braille Points from Braille Dots, (b) Detecting Margin Lines and Start Point ($P_0$) Using Geometric Procedures, (c) Forming Grid Structure from Dot Alignment, (d) Projecting Grid Structure over The Pre-Processed Image for Translation.

The preprocessing phase improves dot formation and removes noise. However, at the same time, it also erases a few numbers of dots (Figure 6).

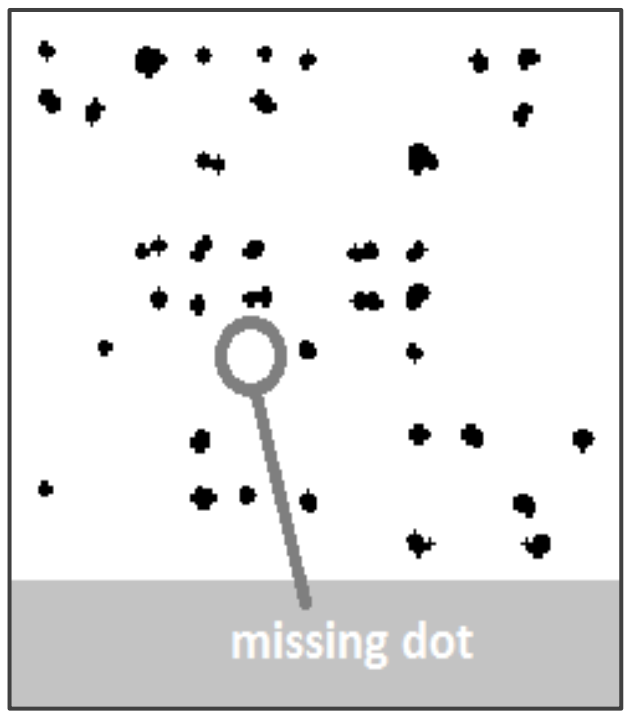

Figure 6: Braille Dot Missing after Preprocessing

*B. Result & Discussion*

The result is evaluated by using 4 metrics- Accuracy Precision, Recall, and F-Measure.

$$\text{Accuracy} = \frac{T_P + T_N}{T_P + T_N + F_P + F_N} \tag{6}$$

$$\text{Precision} = \frac{T_P}{T_P + F_P} \tag{7}$$

$$\text{Recall} = \frac{T_P}{T_P + F_N} \tag{8}$$

$$\text{F} - \text{Measure} = \frac{2 * \text{Prec.} * \text{Rec.}}{\text{Prec.} + \text{Rec.}} \tag{9}$$

Here,

$T_P$ = true-positive

$T_N$ = true-negative

$F_P$ = false-positive

$F_N$ = false-negative

These metrics present a better understanding of the performance of the proposed translation method. Result is shown in Table II below-

TABLE II

EXPERIMENTAL RESULT FOR DOT DETECTION

| $T_P$ | $F_P$ | $T_N$ | $F_N$ | Prec. | Rec. | F-M |
|---|---|---|---|---|---|---|
| 1511 | 0 | 2836 | 21 | 100% | 98.63% | 99.31% |

This high precision rate of dot detection is mainly due to the carefully designed pre-processing phase. In character recognition, we get 97.25% accuracy without performing any post-processing (like- spell checking).

Analysis of the data also unmasks that- noise on the boundary posing as braille dots may source major threat to the performance of the translator.

## V. CONCLUSION

Braille is the key for visually impaired people to enter the realm of world knowledge. So A Robust Braille to Text Translator for all languages will add a significant value to the whole society. The system proposed here is primarily prepared for only keeping Bengali language in mind; however it is divided into separate independent components making it suitable to adapt in different environments. Like- in the translation phase simply altering the mapping (Table 1) will produce text of any other language. The system also does not require any special hardware or configuration for image acquisition. It also confronts with real life complications like- image deterioration, rotation, skew, noise, and deformation. It has been applied to different conditions, and the test result proves its effectiveness and usability. The actual tool also includes some extra algorithms and features for combating different situations. User can also see result produced in each step of the process, and configure for better output.

During this research various problems were faced during data collection as there was no availability of Bengali Braille Writing with text. Also there is no benchmark or standard dataset to test the proposed algorithms. Database of Braille Writing Image and text had to be created as part of the research for measuring the performance and effectiveness of the process.

Still there are great opportunities open for research in this field of OBR. Much work needs to be done on using mobile camera or standard camera for acquisition. Also the word identification accuracy can be immensely improved by incorporating spell checking division in post processing. Natural Language Processing can be attached to the post processing for getting a final boost in result.